\documentclass{article}
\usepackage{spconf}
\usepackage{amsmath,amssymb,mathtools,bm}
\usepackage{graphicx}
\usepackage{booktabs}
\usepackage{multirow}
\usepackage{microtype}
\usepackage{enumitem}
\usepackage{algorithm}
\usepackage{algpseudocode}
\usepackage{siunitx}
\usepackage{placeins}
\usepackage{caption}
\usepackage{xcolor}
\usepackage{makecell}

\def\Title{Task-Agnostic Noisy Label Detection via Standardized Loss Aggregation}
\def\Authors{Inhyuk Park \quad Doohyun Park$^{\dagger}$%
\thanks{$^{\dagger}$Corresponding author: \texttt{doohyun.park@vuno.co}}%
\thanks{Accepted at IEEE ISBI 2026.}}
\def\AffilA{VUNO Inc.}

\title{\Title}
\name{\Authors}
\address{\AffilA}

\begin{document}
\ninept
\maketitle

\begin{abstract}
Noisy labels are common in large-scale medical imaging datasets due to inter-observer variability and ambiguous cases.
We propose a statistically grounded and task-agnostic framework, \textit{Standardized Loss Aggregation} (SLA), for detecting noisy labels at the sample level.
SLA quantifies label reliability by aggregating standardized fold-level validation losses across repeated cross-validation runs.
This formulation generalizes discrete hard-counting schemes into a continuous estimator that captures both the frequency and magnitude of performance deviations, yielding interpretable and statistically stable noisiness scores.
Experiments on a public fundus dataset demonstrate that SLA consistently outperforms the hard-counting baseline across all noise levels and converges substantially faster, especially under low noise ratios where subtle loss variations are informative.
Samples with high SLA scores indicate potentially ambiguous or mislabeled cases, guiding efficient re-annotation and improving dataset reliability for any classification task.
\end{abstract}


\section{Introduction}

Deep learning has achieved remarkable progress in computer vision and medical image analysis. However, its performance critically depends on large-scale datasets with reliable annotations~. In clinical imaging, obtaining such annotations is both costly and challenging: expert labeling is time-consuming and often inconsistent due to inter-observer variability, ambiguous cases, and systematic biases~\cite{shi2024survey, schmidt2023probabilistic, yang2024limits}. Consequently, label noise is common and can adversely affect optimization stability and model generalization~\cite{song2022learning}. Prior studies have shown that noisy supervision biases parameter estimation, slows convergence, and degrades model reliability, underscoring the need for reliable and task-agnostic methods to identify and characterize label noise—especially in domains where re-annotation is costly and heterogeneous uncertainty is prevalent~\cite{shi2024survey, song2022learning, patrini2017making}.

Existing approaches typically aim to learn directly from noisy labels by employing robust loss functions~\cite{ghosh2017robust, wang2019symmetric}, correcting losses using estimated noise transition matrices~\cite{patrini2017making}, or leveraging meta-learning frameworks that adapt training strategies with a small set of trusted clean samples~\cite{wei2024deep}. Approaches such as MentorNet and Co-teaching adopted strategies to separate clean and noisy subsets, improving robustness to label noise~\cite{jiang2018mentornet, han2018co}. While these techniques mitigate the negative effects of label noise, they primarily focus on enhancing model robustness rather than identifying which individual samples are unreliable, limiting interpretability and hindering retrospective dataset curation.

Another line of research aims to explicitly detect noisy labels, such as O2U-Net~\cite{huang2019o2u}, FINE~\cite{kim2021fine}, and repeated cross-validation (ReCoV)~\cite{chen2024detecting}, which identify potentially mislabeled samples. Among these, ReCoV detects noisy samples by repeatedly partitioning the dataset and counting how often each sample appears in the worst-performing fold. Its simplicity and empirical effectiveness make it an appealing baseline for noisy-label detection. However, this discrete, count-based formulation ignores the relative magnitude of fold-wise performance differences and discards distributional information contained in the losses. Consequently, samples in the worst fold receive identical penalties regardless of the actual performance gap, reducing the granularity and statistical interpretability of noisiness estimation.

\FloatBarrier

{\setlength{\abovecaptionskip}{2pt}
 \setlength{\belowcaptionskip}{0pt}
\begin{table}[!t]
  \caption{\textbf{Standardized Loss Aggregation (SLA) for noisy-label detection.}
  Algorithmic procedure for per-sample noisiness estimation via standardized fold-level loss aggregation.}
  \label{tab:sla}
  \centering
  \renewcommand{\arraystretch}{1.1}
  \resizebox{\columnwidth}{!}{%
  \begin{tabular}{p{0.97\linewidth}}
  \toprule
  \textbf{Inputs:} dataset $\{(I_i, y_i)\}_{i=1}^{N}$ with observed labels $Y=\{y_i\}_{i=1}^{N}$,\\
  pretrained encoder $\phi(\cdot)$, number of folds $K$, number of repetitions $R$ \\[1pt]
  \textbf{Output:} per-sample standardized loss scores $\{S_i\}_{i=1}^{N}$ \\[1pt]
  \midrule
  Extract feature embeddings $Z=\{\mathbf{z}_i\}_{i=1}^{N}$ with $\mathbf{z}_i \gets \phi(I_i)$ \\[1pt]
  Fit PCA on $Z$ to obtain a projection matrix $P \in \mathbb{R}^{D \times d}$ (e.g., $d{=}10$); set $X \gets ZP$ \\[1pt]
  \textbf{for each repetition} $r = 1, \dots, R$: \\[1pt]
  \quad Generate stratified $K$-fold split $\{(\mathcal{T}_{r,k}, \mathcal{V}_{r,k})\}_{k=1}^{K}$ \\[1pt]
  \quad \textbf{for each fold} $k = 1, \dots, K$: \\[1pt]
  \qquad Fit classifier $g_{r,k}$ on $(X_{\mathcal{T}_{r,k}}, Y_{\mathcal{T}_{r,k}})$ \\[1pt]
  \qquad Obtain predicted probabilities $p_{r,k}$ on $\mathcal{V}_{r,k}$ \\[1pt]
  \qquad Compute validation loss $\ell_{r,k} = \mathcal{L}_{\mathrm{CE}}\!\big(Y_{\mathcal{V}_{r,k}},\, p_{r,k}\big)$ \\[1pt]
  \quad $\mu_r = \operatorname{mean}_k(\ell_{r,k}),\;\; \sigma_r = \operatorname{std}_k(\ell_{r,k})$ \\[1pt]
  \quad $s_{r,k} = (\ell_{r,k} - \mu_r) / \max(\sigma_r, \varepsilon)$ \\[1pt]
  \quad Update $S_i \gets S_i + s_{r,k}$ for each $i \in \mathcal{V}_{r,k}$ \\[1pt]
  Normalize final scores: $S_i \gets S_i / R$ \\[1pt]
  \textbf{return} $\{S_i\}$ \\[1pt]
  \bottomrule
  \end{tabular}}
\end{table}
}

To address these limitations, we propose \textit{Standardized Loss Aggregation} (SLA)—a statistically grounded and task-agnostic framework that aggregates standardized fold-level validation losses across repeated cross-validation runs to derive continuous noisiness scores for individual samples. This continuous formulation generalizes the discrete hard-counting scheme into a smooth estimator that captures both the frequency and magnitude of underperformance, yielding interpretable, reproducible, and model-independent label reliability scores.

The main contributions of this work are as follows:
(1) We introduce SLA, a standardized aggregation mechanism that quantifies per-sample label noisiness through continuous fold-level loss integration.
(2) We empirically demonstrate that SLA consistently outperforms discrete count-based baselines in detecting ambiguous or mislabeled samples across varying noise levels, achieving faster and more stable convergence.

\begin{figure*}[t]
\centering
\includegraphics[width=\textwidth]{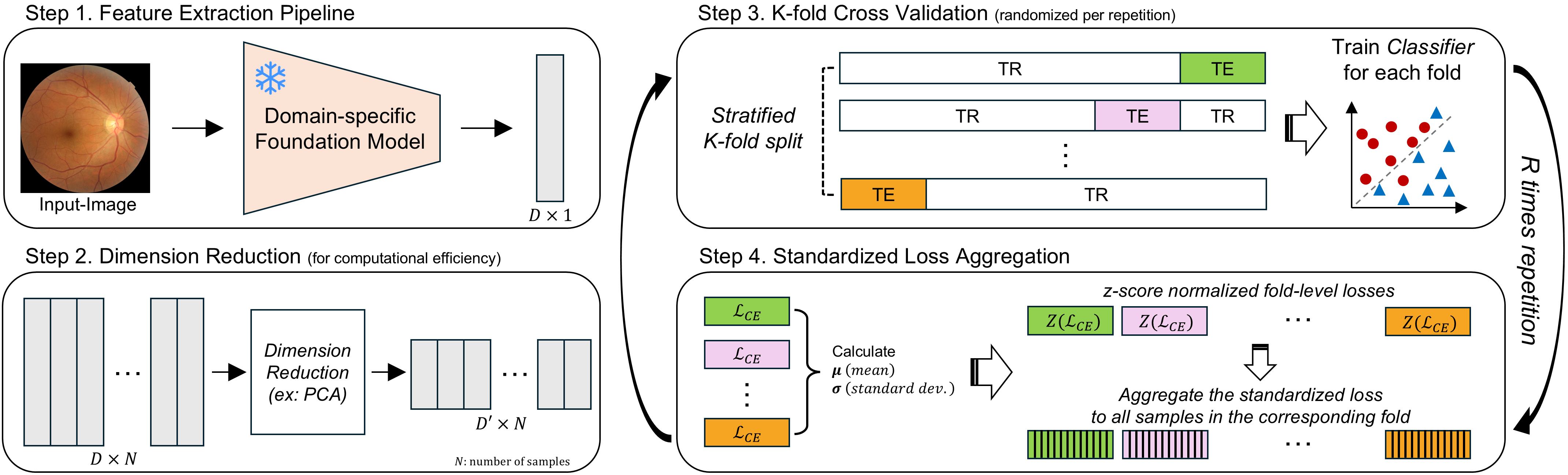}
\caption{Overview of the proposed Standardized Loss Aggregation (SLA) pipeline. 
Repeated $K$-fold cross validation produces fold-level losses that are standardized within each run 
and aggregated per sample to obtain continuous noisiness scores.}
\label{fig:pipeline}
\end{figure*}

\begin{figure*}[t]
\centering
\includegraphics[width=\textwidth]{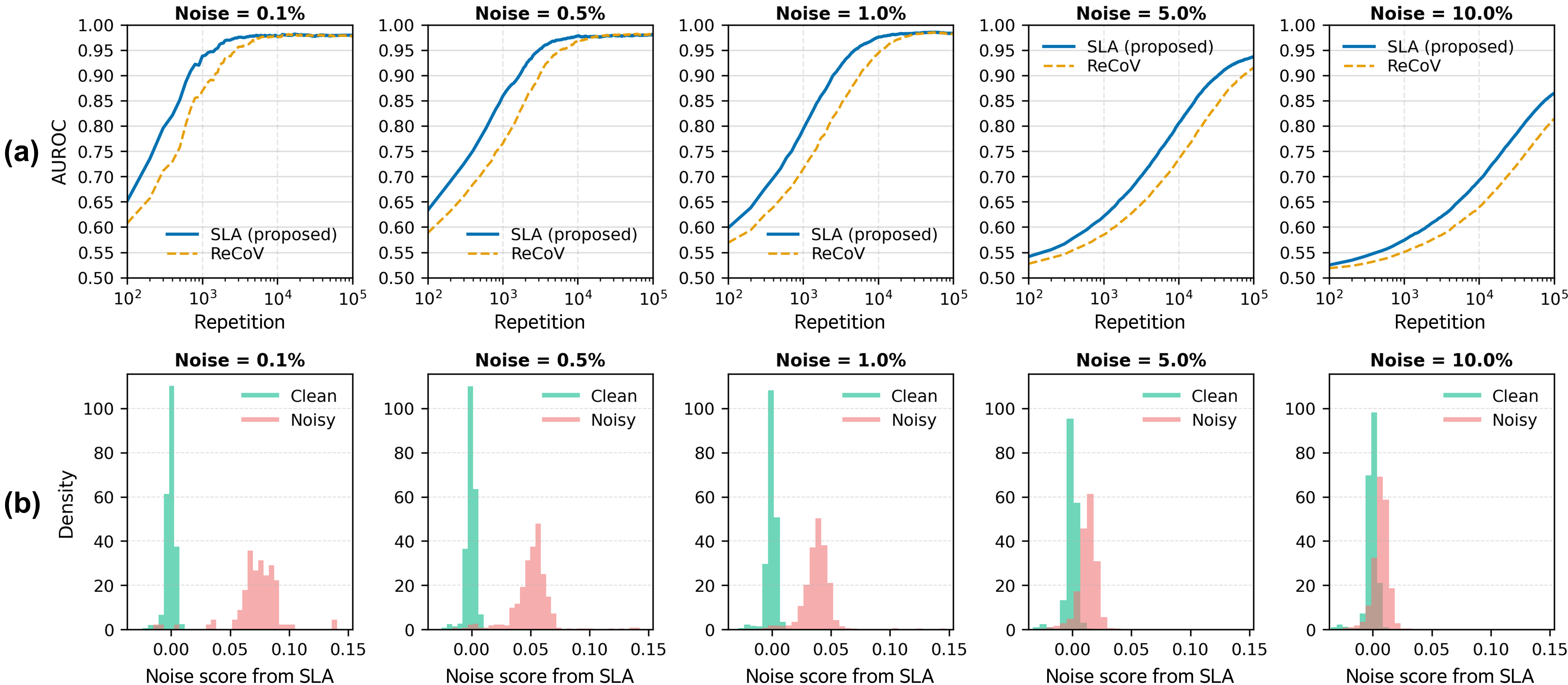}
\caption{
\textbf{Performance and score analysis of the proposed SLA method.}
\textbf{(a)}~AUROC comparison across repetitions
at label-noise ratios of 0.1\%, 0.5\%, 1\%, 5\%, and 10\%.
SLA (proposed) consistently outperforms ReCoV during early repetitions,
and their performance gradually converges as more repetitions are aggregated.
\textbf{(b)}~Standardized-loss score distributions for clean and noisy samples under SLA
after 100k repetitions.
Clean samples are centered around zero, whereas noisy samples form a separate distribution
with a higher mean, indicating larger average loss magnitudes for mislabeled data.
}
\label{fig:sla_combined}
\end{figure*}

\section{Method}
\label{sec:method}

\subsection{Overview}
The proposed framework, called \emph{Standardized Loss Aggregation (SLA)}, estimates sample-level label noisiness by aggregating standardized fold-level validation losses across repeated $K$-fold cross-validation.  
Unlike discrete count-based methods that only record occurrences in the worst-performing folds, SLA continuously integrates normalized loss deviations, yielding a statistically grounded, task-agnostic, and fine-grained measure of label reliability. Figure~\ref{fig:pipeline} illustrates the overall process,
and Table~\ref{tab:sla} summarizes the algorithm.

\subsection{Standardized Loss Aggregation framework}
For each repetition $r = 1, \dots, R$, a stratified $K$-fold partition $\{(\mathcal{T}_{r,k}, \mathcal{V}_{r,k})\}_{k=1}^{K}$ is generated.  
A classifier $g_{r,k}$ is trained on $\mathcal{T}_{r,k}$ and evaluated on $\mathcal{V}_{r,k}$, yielding a fold-level validation loss:
\begin{equation}
\ell_{r,k} = \frac{1}{|\mathcal{V}_{r,k}|} 
\sum_{i \in \mathcal{V}_{r,k}} 
\Big[-y_i \log \hat{p}_{r,k}(i) - (1 - y_i)\log(1 - \hat{p}_{r,k}(i))\Big].
\label{eq:foldloss}
\end{equation}

Within each repetition, fold-level losses are standardized to remove scale differences:
\begin{equation}
\mu_r = \operatorname{mean}_k(\ell_{r,k}), \quad
\sigma_r = \operatorname{std}_k(\ell_{r,k}),
\end{equation}
\begin{equation}
s_{r,k} = \frac{\ell_{r,k} - \mu_r}{\max(\sigma_r, \varepsilon)}.
\label{eq:zscore}
\end{equation}

Each sample $i$ inherits the standardized loss $s_{r,k}$ corresponding to the validation subset $\mathcal{V}_{r,k}$ to which it belongs.  
Aggregating across all repetitions yields the final per-sample noisiness score:
\begin{equation}
S_i = \frac{1}{R} \sum_{r=1}^{R} s_{r,\,k(i;r)},
\label{eq:score}
\end{equation}
where $k(i;r)$ denotes the validation fold containing sample $i$ in repetition $r$.  
This continuous score reflects both the consistency and magnitude of fold-level underperformance across randomized splits, providing a stable and interpretable indicator of sample-level label reliability.

\subsection{Dataset and preprocessing}
Experiments were conducted on the \emph{Justified Referral in AI Glaucoma Screening (JustRAIGS)} dataset~\cite{madadi2025justraigs},  
a publicly available collection of retinal fundus photographs for glaucoma screening.  
The dataset comprises 101{,}423 images in total, including 3{,}270 positive (\textit{referable glaucoma}, RG) and 98{,}153 negative (\textit{non-referable glaucoma}, NRG) cases.  

Each image was independently annotated by two graders (G1 and G2) randomly selected from a pool of qualified ophthalmic graders.  
If both graders agreed on the classification (RG or NRG), their decision was taken as the final label.  
In cases of disagreement, a third grader (G3)—a board-certified glaucoma specialist—reviewed the image,  
and their assessment was used as the consensus label.  
Despite this multi-grader protocol, inter-observer variability and borderline presentations introduce intrinsic uncertainty,  
making the dataset well suited for evaluating methods that detect or quantify label noise.  

All images were resized to $512\times512$ pixels and channel-wise normalized before feature extraction.  
To obtain compact and semantically rich representations,  
we used the FLAIR (Fundus--Language--Aligned Image Representation) foundation model~\cite{silva2025foundation}  
as a frozen feature extractor $\phi(\cdot)$.  
FLAIR is a large-scale vision--language pretrained model for ophthalmic imaging,  
trained on millions of fundus--report pairs to capture pathology-relevant features.  
For each image $I_i$, we extracted embeddings $\mathbf{z}_i = \phi(I_i) \in \mathbb{R}^{D}$, where $D = 2048$.  
The encoder was kept frozen to ensure that noisiness reflects label-level uncertainty rather than representation instability.

\subsection{Statistical perspective and lightweight implementation}
The SLA output can be interpreted as a large-sample statistical estimator obtained through repeated random partitioning.
As defined in Eq.~\ref{eq:score}, each sample $i$ has a final score $S_i$.  
Under mild regularity conditions such as approximate independence of repetitions and finite variance of standardized losses,  
$S_i$ serves as an unbiased estimator of $\mathbb{E}[s_{r,\,k(i;r)}]$,  
with sampling variance decreasing as $\mathrm{Var}(S_i)=\mathrm{Var}(s_{r,\,k(i;r)})/R$.  
By the law of large numbers, $S_i$ converges in probability to its expectation as $R \to \infty$,  
providing a statistically grounded estimate of per-sample noisiness rather than a heuristic aggregation.

Performing $R$ repetitions of $K$-fold validation requires training $R \times K$ classifiers,  
which can be computationally demanding.  
To ensure tractability, SLA was designed to be lightweight and reproducible.  
Specifically:  
(i) the pretrained encoder $\phi(\cdot)$ is fixed so that embeddings are computed only once;  
(ii) the extracted features are reduced using PCA, yielding $X = \mathrm{PCA}(Z) \in \mathbb{R}^{N \times 10}$,  
which substantially reduces the per-fold training cost; and  
(iii) a simple probabilistic classifier is adopted for efficient repetition at large $R$.

We employ linear discriminant analysis (LDA) as the base classifier $g_\theta : \mathbb{R}^{d} \rightarrow [0,1]$,  
which performs the dataset-specific prediction task (e.g., referable vs.\ non-referable classification on JustRAIGS)  
and produces class posterior probabilities $\hat{p}_i = g_\theta(x_i)$.  
LDA avoids iterative optimization and provides stable probabilistic outputs,  
allowing hundreds or even thousands of cross-validation repetitions to be executed efficiently.

\subsection{Interpretation and relation to discrete baselines}
The standardized aggregation normalizes fold-level losses within each repetition,  
allowing direct comparison across validations despite differences in overall run difficulty.  
Instead of measuring absolute task performance, SLA captures the relative deviation of each sample's loss from its fold mean.  
Accordingly, $S_i$ quantifies how consistently a sample exhibits higher losses than its peers.  
Averaging across randomized splits further stabilizes the estimates and suppresses sampling noise.

This formulation generalizes discrete fold-count methods such as ReCoV~\cite{chen2024detecting}.  
While ReCoV assigns identical penalties to all samples appearing in the worst-performing fold,  
SLA aggregates standardized losses continuously, preserving information about the magnitude of deviation.  
As a result, SLA produces smoother and statistically interpretable noisiness scores  
that remain robust to random fold assignments and dataset heterogeneity.

\section{Results}
\label{sec:results}

\subsection{Controlled label-noise experiments}
We evaluated the proposed framework under controlled synthetic conditions using the \emph{JustRAIGS} dataset~\cite{madadi2025justraigs},  
where random label flips were applied to simulate noise ratios ranging from 0.1\% to 10\%.  
Each experiment was conducted with $K=5$ folds and $R=100{,}000$ repetitions.  

Figure~\ref{fig:sla_combined} summarizes the area under the receiver operating characteristic curve (AUROC) results for noisy-sample detection across different noise ratios.  
Across all noise levels, the proposed \textit{soft scoring} approach (SLA) consistently achieved higher AUROC values than the discrete \textit{hard counting} baseline (ReCoV).  
As shown in Figure~\ref{fig:sla_combined}(a), SLA exhibits a clear advantage during the early repetition stages,  
while the two methods gradually converge in performance as the number of repetitions increases.  

Table~\ref{tab:sim-auroc} provides a quantitative comparison after 1{,}000 repetitions,  
showing that SLA yields consistently higher AUROC values across all noise ratios.  
The improvement is particularly pronounced at lower noise levels.  
For instance, at a 0.5\% noise ratio, AUROC increased from 0.765 with ReCoV to 0.859 with SLA,  
demonstrating that the proposed method offers greater robustness to small fractions of label corruption and converges faster during early repetitions.

Figure~\ref{fig:sla_combined}(b) presents the standardized-loss score distributions under SLA after 100{,}000 repetitions.  
Clean samples are centered near zero, whereas noisy samples form a distinct distribution with a higher mean.  
As the noise ratio increases, the overlap between these two distributions becomes larger,  
indicating reduced discriminability between clean and mislabeled data at the same repetition. 
Consistent with Figure~\ref{fig:sla_combined}(a), AUROC at higher noise levels (e.g., 5.0\% and 10.0\%) have not yet reached saturation by 100{,}000 repetitions, suggesting that additional repetitions may further improve separation.  
This observation implies that, for datasets with higher expected noise,  
the optimal number of repetitions can be adaptively determined by monitoring the per-repetition AUROC trend— 
analogous to applying an \textit{early stopping} criterion in conventional deep learning training.

\subsection{Normalized interpretability and task-agnostic behavior}
Empirical results from controlled noise experiments confirm the interpretability and normalization properties of SLA.  
Because the aggregated scores are dimensionless and normalized by $R$,  
their magnitude remains invariant to the number of repetitions and consistent across datasets and architectures.  
This invariance allows $\{S_i\}$ to serve as a unified reliability index,  
facilitating direct comparison of label quality across models and domains.  
Furthermore, because SLA depends solely on validation losses,  
it operates in a fully task-agnostic manner and can be seamlessly applied to any classification pipeline  
without requiring model retraining or architectural modification.

{\setlength{\abovecaptionskip}{2pt}
 \setlength{\belowcaptionskip}{0pt}
\begin{table}[t]
\centering
\caption{Noisy-label sample detection performance (AUROC) across different noise ratios after 1{,}000 repetitions. 
Red text indicates improvement over the baseline.}
\label{tab:sim-auroc}
\setlength{\tabcolsep}{7pt}
\renewcommand{\arraystretch}{1.0}
\begin{tabular}{rccccc}  
\toprule
\multicolumn{1}{c}{\smash{\raisebox{-8pt}{\textbf{Method}}}} &  
\multicolumn{5}{c}{\textbf{Noise ratio}} \\
\cmidrule(l{1pt}r{1pt}){2-6}
& 0.1\% & 0.5\% & 1.0\% & 5.0\% & 10.0\% \\
\midrule
ReCoV & 0.869 & 0.765 & 0.715 & 0.585 & 0.551 \\
\makecell[r]{SLA\\[-2pt]{\scriptsize(proposed)}} &
\makecell[c]{0.938\\[-2pt]{\scriptsize\textcolor{red}{(+0.069)}}} &
\makecell[c]{0.859\\[-2pt]{\scriptsize\textcolor{red}{(+0.094)}}} &
\makecell[c]{0.793\\[-2pt]{\scriptsize\textcolor{red}{(+0.078)}}} &
\makecell[c]{0.621\\[-2pt]{\scriptsize\textcolor{red}{(+0.036)}}} &
\makecell[c]{0.574\\[-2pt]{\scriptsize\textcolor{red}{(+0.023)}}} \\
\bottomrule
\end{tabular}
\end{table}
}

\subsection{Computational efficiency and convergence behavior}

All computations were performed on a Linux workstation (Ubuntu 22.04 LTS) 
equipped with dual Intel Xeon E5-2620~v4 CPUs (2.10~GHz, 32~threads) and 128~GB of RAM, without GPU acceleration.  
On average, one SLA repetition---including five-fold training and validation---required approximately 0.9~seconds on CPU.

As shown in Figure~\ref{fig:sla_combined}(a), the proposed method exhibited rapid convergence of AUROC across all simulated noise ratios.  
For low noise levels (e.g., 0.1\%--1.0\%), SLA reached near-saturation performance around 5{,}000~repetitions, 
corresponding to approximately 75~minutes of computation.  
At higher noise ratios (5.0\%--10.0\%), the AUROC continued to improve beyond 100{,}000~repetitions, 
indicating that full convergence was not yet achieved under severe noise conditions.

\section{Discussion}
\label{sec:discussion}

The proposed \emph{Standardized Loss Aggregation (SLA)} framework provides a statistically grounded and task-agnostic approach for detecting noisy labels at the sample level.  
By aggregating standardized fold-level validation losses across repeated cross-validation runs,  
SLA quantifies label reliability as the expected standardized deviation of each sample’s loss from the fold mean.  
This formulation generalizes discrete hard-counting schemes~\cite{chen2024detecting} and relates conceptually to confidence-based noisy-label estimation frameworks~\cite{northcutt2021confident},  
yielding interpretable, reproducible, and model-independent noisiness scores.

Empirically, SLA consistently outperformed the hard-counting baseline across all simulated noise levels,
with the largest gains observed at lower noise ratios where subtle loss variations remain informative.
It achieved higher performance in early repetitions and reached saturation substantially faster,
whereas the baseline improved more gradually before eventual convergence.
This contrast reflects their aggregation mechanisms:
SLA continuously integrates standardized losses from all folds,
whereas ReCoV updates scores only for samples in the worst-performing fold, resulting in a coarser and temporally sparse aggregation process.
This behavior aligns with the statistical foundation of ensemble-based evidence accumulation~\cite{dietterich2000ensemble},
where continuous aggregation operates as an averaging estimator over partially independent folds, reducing estimator variance and mitigating selection bias.
Under this view, SLA can be interpreted as computing a cross-fold Monte Carlo approximation of the expected standardized deviation,
yielding a consistent and asymptotically lower-variance estimator of sample-level noisiness.

The standardized-loss distributions further support these findings.  
Clean samples cluster near zero, whereas mislabeled samples exhibit consistently positive deviations,  
confirming that SLA separates reliable from unreliable data on a continuous scale.  
Because the scores are normalized and dimensionless, they remain directly comparable across datasets, architectures, and repetition counts,  
enabling consistent interpretation and robust label quality assessment across diverse classification tasks.  
These properties position SLA as a generalizable framework for disentangling label corruption from intrinsic data uncertainty, consistent with mixture-modeling perspectives in noisy-label learning~\cite{song2022learning, li2020dividemix}.

Mechanistically, SLA links fold-level loss dynamics to sample-specific reliability.  
Mislabeled or ambiguous samples systematically inflate fold losses, producing persistent positive standardized deviations after normalization.  
Aggregating these deviations across randomized partitions captures both label corruption and intrinsic ambiguity,  
providing a unified signal of per-sample uncertainty.

From a practical standpoint, SLA is computationally efficient owing to its lightweight design---fixed feature extraction,  
PCA-based dimensionality reduction, and repeated training of linear discriminant classifiers.  
Despite operating solely on CPUs, the full computation completed within roughly one day even for large-scale experiments exceeding 100{,}000 fundus images,  
demonstrating that reliable noisy-label estimation can be achieved without GPU acceleration.  
Such efficiency underscores the practical scalability of SLA, enabling rapid dataset auditing and convergence analysis in real-world research environments  
where computational resources may be limited.  
In medical imaging applications, where annotations are often obtained from multiple graders and finalized through consensus,  
residual disagreement and borderline cases frequently lead to residual noisy labels even after consensus.  
In such contexts, the continuous noisiness scores produced by SLA can serve as a quantitative aid for selecting samples  
that warrant re-evaluation or additional expert review.  
By prioritizing high score samples for re-annotation, practitioners can focus expert effort on the most uncertain cases,  
reducing redundant manual review while improving dataset reliability.  
This targeted refinement supports transparent and efficient data governance, particularly in clinical domains where annotation resources are limited.

Despite these advantages, several limitations remain.  
First, our evaluation relies on synthetic noise introduced via random label flips,  
which isolates methodological effects but does not fully represent real-world annotation variability.  
Second, high noiseness scores may also correspond to inherently ambiguous or borderline samples,  
rather than truly incorrect labels.  
Finally, while SLA produces a continuous reliability spectrum,  
determining a universal threshold for label filtering remains task-dependent and somewhat arbitrary.

In summary, SLA offers a principled, computationally efficient, and statistically interpretable framework for noisy-label detection.  
Its convergence properties, normalization-based interpretability, and scalability establish a versatile foundation  
for quantitative label quality assessment and systematic dataset refinement.

\section{COMPLIANCE WITH ETHICAL STANDARDS}
This research study was conducted retrospectively using human subject data made available in open access. Ethical approval was not required as confirmed by the license attached with the open access data.

\section{Conflict of Interest}
The authors are employees of VUNO Inc., but declare that they have no competing financial or non-financial interests related to this work.

\bibliographystyle{IEEEtran}
\bibliography{refs}

@article{madadi2025justraigs,
  title={JustRAIGS: Justified Referral in AI Glaucoma Screening Challenge},
  author={Madadi, Yeganeh and Raja, Hina and Vermeer, Koenraad A and Lemij, Hans G and Huang, Xiaoqin and Kim, Eunjin and Lee, Seunghoon and Kwon, Gitaek and Kim, Hyunwoo and Kim, Jaeyoung and others},
  journal={IEEE Transactions on Medical Imaging},
  year={2025},
  publisher={IEEE}
}

@article{silva2025foundation,
  title={A foundation language-image model of the retina (flair): Encoding expert knowledge in text supervision},
  author={Silva-Rodriguez, Julio and Chakor, Hadi and Kobbi, Riadh and Dolz, Jose and Ayed, Ismail Ben},
  journal={Medical Image Analysis},
  volume={99},
  pages={103357},
  year={2025},
  publisher={Elsevier}
}

@inproceedings{patrini2017making,
  title={Making deep neural networks robust to label noise: A loss correction approach},
  author={Patrini, Giorgio and Rozza, Alessandro and Krishna Menon, Aditya and Nock, Richard and Qu, Lizhen},
  booktitle={Proceedings of the IEEE conference on computer vision and pattern recognition},
  pages={1944--1952},
  year={2017}
}

@inproceedings{ghosh2017robust,
  title={Robust loss functions under label noise for deep neural networks},
  author={Ghosh, Aritra and Kumar, Himanshu and Sastry, P Shanti},
  booktitle={Proceedings of the AAAI conference on artificial intelligence},
  volume={31},
  number={1},
  year={2017}
}

@inproceedings{wang2019symmetric,
  title={Symmetric cross entropy for robust learning with noisy labels},
  author={Wang, Yisen and Ma, Xingjun and Chen, Zaiyi and Luo, Yuan and Yi, Jinfeng and Bailey, James},
  booktitle={Proceedings of the IEEE/CVF international conference on computer vision},
  pages={322--330},
  year={2019}
}

@inproceedings{jiang2018mentornet,
  title={Mentornet: Learning data-driven curriculum for very deep neural networks on corrupted labels},
  author={Jiang, Lu and Zhou, Zhengyuan and Leung, Thomas and Li, Li-Jia and Fei-Fei, Li},
  booktitle={International conference on machine learning},
  pages={2304--2313},
  year={2018},
  organization={PMLR}
}

@article{han2018co,
  title={Co-teaching: Robust training of deep neural networks with extremely noisy labels},
  author={Han, Bo and Yao, Quanming and Yu, Xingrui and Niu, Gang and Xu, Miao and Hu, Weihua and Tsang, Ivor and Sugiyama, Masashi},
  journal={Advances in neural information processing systems},
  volume={31},
  year={2018}
}

@inproceedings{chen2024detecting,
  title={Detecting Noisy Labels with Repeated Cross-Validations},
  author={Chen, Jianan and Ramanathan, Vishwesh and Xu, Tony and Martel, Anne L},
  booktitle={International Conference on Medical Image Computing and Computer-Assisted Intervention},
  pages={197--207},
  year={2024},
  organization={Springer}
}

@article{wei2024deep,
  title={Deep learning with noisy labels in medical prediction problems: a scoping review},
  author={Wei, Yishu and Deng, Yu and Sun, Cong and Lin, Mingquan and Jiang, Hongmei and Peng, Yifan},
  journal={Journal of the American Medical Informatics Association},
  volume={31},
  number={7},
  pages={1596--1607},
  year={2024},
  publisher={Oxford Academic}
}

@article{shi2024survey,
  title={A survey of label-noise deep learning for medical image analysis},
  author={Shi, Jialin and Zhang, Kailai and Guo, Chenyi and Yang, Youquan and Xu, Yali and Wu, Ji},
  journal={Medical image analysis},
  volume={95},
  pages={103166},
  year={2024},
  publisher={Elsevier}
}

@inproceedings{schmidt2023probabilistic,
  title={Probabilistic modeling of inter-and intra-observer variability in medical image segmentation},
  author={Schmidt, Arne and Morales-Alvarez, Pablo and Molina, Rafael},
  booktitle={Proceedings of the IEEE/CVF international conference on computer vision},
  pages={21097--21106},
  year={2023}
}

@article{yang2024limits,
  title={The limits of fair medical imaging AI in real-world generalization},
  author={Yang, Yuzhe and Zhang, Haoran and Gichoya, Judy W and Katabi, Dina and Ghassemi, Marzyeh},
  journal={Nature Medicine},
  volume={30},
  number={10},
  pages={2838--2848},
  year={2024},
  publisher={Nature Publishing Group US New York}
}

@article{song2022learning,
  title={Learning from noisy labels with deep neural networks: A survey},
  author={Song, Hwanjun and Kim, Minseok and Park, Dongmin and Shin, Yooju and Lee, Jae-Gil},
  journal={IEEE transactions on neural networks and learning systems},
  volume={34},
  number={11},
  pages={8135--8153},
  year={2022},
  publisher={IEEE}
}

@inproceedings{li2020dividemix,
  title={DivideMix: Learning with Noisy Labels as Semi-supervised Learning},
  author={Li, Junnan and Socher, Richard and Hoi, Steven CH},
  booktitle={ICLR},
  year={2020}
}

@article{northcutt2021confident,
  title={Confident learning: Estimating uncertainty in dataset labels},
  author={Northcutt, Curtis and Jiang, Lu and Chuang, Isaac},
  journal={Journal of Artificial Intelligence Research},
  volume={70},
  pages={1373--1411},
  year={2021}
}

@inproceedings{dietterich2000ensemble,
  title={Ensemble methods in machine learning},
  author={Dietterich, Thomas G},
  booktitle={International workshop on multiple classifier systems},
  pages={1--15},
  year={2000},
  organization={Springer}
}

@inproceedings{huang2019o2u,
  title={O2u-net: A simple noisy label detection approach for deep neural networks},
  author={Huang, Jinchi and Qu, Lie and Jia, Rongfei and Zhao, Binqiang},
  booktitle={Proceedings of the IEEE/CVF international conference on computer vision},
  pages={3326--3334},
  year={2019}
}

@article{kim2021fine,
  title={Fine samples for learning with noisy labels},
  author={Kim, Taehyeon and Ko, Jongwoo and Choi, JinHwan and Yun, Se-Young and others},
  journal={Advances in Neural Information Processing Systems},
  volume={34},
  pages={24137--24149},
  year={2021}
}
\end{document}